\documentclass[lettersize,journal]{IEEEtran}
\usepackage{amsmath,amsfonts}
\usepackage{algorithmic}
\usepackage{algorithm}
\usepackage{array}
\usepackage[caption=false,font=normalsize,labelfont=sf,textfont=sf]{subfig}
\usepackage{textcomp}
\usepackage{stfloats}
\usepackage{url}
\usepackage{verbatim}
\usepackage{graphicx}
\usepackage{cite}
\def\BibTeX{{\rm B\kern-.05em{\sc i\kern-.025em b}\kern-.08em
    T\kern-.1667em\lower.7ex\hbox{E}\kern-.125emX}}
\usepackage{balance}

\usepackage[numbers,sort&compress]{natbib}
\usepackage[table]{xcolor} 
\usepackage{booktabs}
\usepackage{multirow}
\usepackage[pagebackref=true,breaklinks=true,colorlinks,bookmarks=True]{hyperref}

\begin{document}

\title{Visual and Textual Prior Guided Mask Assemble for Few-Shot Segmentation and Beyond}

\author{Shuai Chen, Fanman Meng, Runtong Zhang, Heqian Qiu, Hongliang Li, Qingbo Wu, Linfeng Xu
\thanks{The Authors are with the School of Information and Communication Engineering, University of Electronic Science and Technology of China, Chengdu, 611731, China (e-mail: schen@std.uestc.edu.cn; fmmeng@uestc.edu.cn; 202211012322@std.uestc.edu.cn; hqqiu@std.uestc.edu.cn; hlli@uestc.edu.cn; qbwu@uestc.edu.cn; lfxu@uestc.edu.cn)}
\thanks{Corresponding author: Fanman Meng.}
}



\maketitle

\begin{abstract}
    Few-shot segmentation (FSS) aims to segment the novel classes with a few annotated images. Due to CLIP's advantages of aligning visual and textual information, the integration of CLIP can enhance the generalization ability of FSS model. However, even with the CLIP model, the existing CLIP-based FSS methods are still subject to the biased prediction towards base classes, which is caused by the class-specific feature level interactions. To solve this issue, we propose a visual and textual Prior Guided Mask Assemble Network (PGMA-Net). It employs a class-agnostic mask assembly process to alleviate the bias, and formulates diverse tasks into a unified manner by assembling the prior through affinity. Specifically, the class-relevant textual and visual features are first transformed to class-agnostic prior in the form of probability map. Then, a Prior-Guided Mask Assemble Module (PGMAM) including multiple General Assemble Units (GAUs) is introduced. It considers diverse and plug-and-play interactions, such as visual-textual, inter- and intra-image, training-free, and high-order ones. Lastly, to ensure the class-agnostic ability, a Hierarchical Decoder with Channel-Drop Mechanism (HDCDM) is proposed to flexibly exploit the assembled masks and low-level features, without relying on any class-specific information. It achieves new state-of-the-art results in the FSS task, with mIoU of $77.6$ on $\text{PASCAL-}5^i$ and $59.4$ on $\text{COCO-}20^i$ in 1-shot scenario. Beyond this, we show that without extra re-training, the proposed PGMA-Net can solve bbox-level and cross-domain FSS, co-segmentation, zero-shot segmentation (ZSS) tasks, leading an any-shot segmentation framework. 
\end{abstract}


\section{Introduction}

\IEEEPARstart{T}{he} remarkable success has been made in the area of semantic segmentation by deep learning based methods~\cite{long2015fully,chen2017deeplab,cheng2021per,QiuYM18,9745353}. However, this progress heavily relies on the availability of large annotated datasets~\cite{lin2014microsoft,everingham2010pascal,zhou2019semantic}, requiring a time-consuming and laborious process of annotation. Additionally, these methods fail to handle the extremely low-data scenario on the novel classes in practice. Conversely, humans are capable of identifying and segmenting novel concepts with a few visual stimulation. Motivated by this, researchers have proposed the few-shot segmentation (FSS) task, aiming to build a class-agnostic model from the base classes, and segment object of any novel classes.

\begin{figure}[t]
    \centering
    \includegraphics[width=\linewidth]{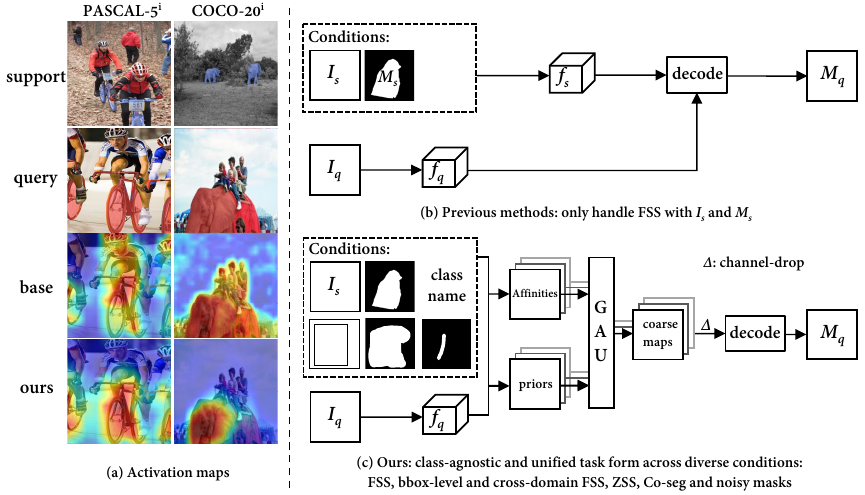}
    \caption{(a): Activation maps on  $\text{PASCAL-}5^i$ and $\text{COCO-}20^i$. Despite incorporating visual information from support as assistance, the baseline struggles to activate the "Bicycle" area and is prone to base class "Person". This is due to the insufficient guidance and an improper utilization of class-specific feature-to-mask mapping. (b): Previous methods is valid solely when both the support image $I_{s}$ and support mask $M_{s}$ are available. (c): This paper proposes to incorporate textual components via a class-agnostic prior-to-mask mapping to address these issues, and is capable of performing diverse tasks (FSS, bbox-level and cross-domain FSS, ZSS, co-segmentation, FSS with inaccurate support mask) in a unified form: assembling the prior by affinity.}
    \label{fig:motivation}
  \end{figure}

However, even equipped with the powerful meta-learning and metric-learning schemes, existing FSS models~\cite{zhang2019pyramid,liu2020part,tian2020prior,boudiaf2021few,min2021hypercorrelation,fan2022self,liu2023fecanet,9999056} still suffer from the inaccurate localization of the target object, along with an over-fitting to base class as shown in Figure~\ref{fig:motivation} (a). We contend that this is originating from two reasons: 1) They rely solely on a few visual annotations, with limitation on handling the vast variations of appearance, geometric and context among objects. 2) They mainly focus on learning a mapping from class-specific features to masks, which makes the mapping biased to the semantically rich features of the base class, leading to ineffectual inference on novel class. Moreover, as shown in Figure~\ref{fig:motivation} (b, c), existing FSS models fail to handle extra forms of guided segmentation tasks via one suite of model weights, such as bounding-box guided FSS, cross-domain FSS, co-segmentation and text-guided zero-shot segmentation tasks.  

This paper devotes to design a \textbf{purely class-agnostic} model to alleviate the training bias when incorporating the semantically rich textual information, and further capable of executing the above tasks in a \textbf{unified task form}. To achieve these goals, we integrate these tasks into a unified formulation that assembles the prior via affinity. On one hand, this formulation takes only the class-agnostic prior and affinity as direct inputs, both of which are class-agnostic. Therefore, it can alleviate training bias compared to the previous class-specific feature-level mapping. On the other hand, different tasks have different priors and affinities. For example, the ZSS task involves intra-image affinity and textual prior, while the FSS task includes inter-image affinity and visual prior. Despite these variations, the interaction mechanism remains the same. This uniformity in such formulation allows us to perform diverse tasks using same set of model weights.

Following the proposed scheme, we propose our PGMA-Net, it includes three main modules: 1) a \textbf{Prior Adapter} that converts class-relevant visual-textual features of CLIP~\cite{radford2021learning} into class-agnostic priors in the form of probability map. Additionally, an \textbf{Affinity Extractor} is also proposed to capture the class-independent pixel-to-pixel correspondence via inter- and intra-image, training-free, and high-order correlations. 2) a Prior-Guided Mask Assemble Module (\textbf{PGMAM}) including multiple General Assemble Units (GAUs) is introduced. It assembles diverse priors by corresponding affinities in a unififed manner. 3) To convert the assembled priors into the final prediction mask in class-agnostic manner, we propose a Hierarchical Decoder with Channel-Drop Mechanism \textbf{(HDCDM)}. It not only essentially ensures the generalization ability by taking in only the assembled priors and low-level features, but also allows our model to perform extra segmentation tasks. Overall, our contributions are as follows:

\begin{enumerate}
    \item We present a new architecture called the Prior Guided Mask Assemble Network (PGMA-Net) for the few-shot segmentation task. This network incorporates textual information by utilizing a class-agnostic prior-to-mask assembly process, which helps to alleviate the training bias towards the base class observed in previous methods.
    \item We introduce Prior-Guided Mask Assemble Module (PGMAM), which formulates diverse tasks in a unified manner by assembling the prior through affinity. It considers diverse plug-and-play interactions between priors (including visual and textual, support and query ones) and affinities (such as inter- and intra-image, training-free, and high-order ones).  
    \item We employ Hierarchical Decoder with Channel-Drop Mechanism (HDCDM), allowing for handling diverse tasks using one suit of mode weights.
    \item We achieve new state-of-the-art results in the FSS task, with mIoU of $77.6$ on $\text{PASCAL-}5^i$ and $59.4$ on $\text{COCO-}20^i$. Moreover, without re-training, the trained PGMA-Net shows promising performance across various tasks, including ZSS, box-level and cross-domain FSS, co-segmentation, FSS with inaccurate support annotations. This unified framework constitutes an any-shot segmentation model.
  \end{enumerate}

\section{Related Work}
\subsection{Few-shot Learning}
Few-shot learning (FSL) intends to construct a classification model for new task with a few labeled examples. Meta-learning is the predominant paradigm for FSL, and it is further categorized into metric-based, optimization-based, and model-based methods. Metric-based methods employ either an embedding network~\cite{vinyals2016matching,snell2017prototypical,ZhangLK23} or a learnable distance~\cite{sung2018learning}. Optimization-based methods aim to learn a well-performed model initialization, followed by quick adaptation~\cite{finn2017model,rajeswaran2019meta}. Model-based methods are designed to create a model structure~\cite{cai2018memory,YangHLHW23} that is specifically tailored to meta-learning. Transfer learning is another way in FSL~\cite{hu2022pushing,zhang2022tip}, where knowledge is transferred from either the base class~\cite{hu2022squeezing}, or pre-trained models like DINO~\cite{caron2021emerging} and CLIP~\cite{radford2021learning}.

\subsection{Few-shot Segmentation}
Few-shot segmentation aims to segment the novel classes with just a few annotated images. Episodic training strategy~\cite{vinyals2016matching} has been employed by previous approaches to learn a class-agnostic model, which  can be subdivided into prototype-based and matching-based methods. The prototype-based methods compressed the support features into class-specific prototypes, and then perform segmentation via fixed cosine distance or a learnable metric. Diverse prototypes can be formed, e.g., a single global foreground prototype obtained through masked average pooling~\cite{snell2017prototypical,wang2019panet,10109193}, multiple foreground prototypes obtained through clustering~\cite{li2021adaptive} and EM~\cite{yang2020prototype}, learnable meta-memory prototypes~\cite{wu2021learning}, and prototypes of base classes~\cite{cheng2023hpa}. However, compressing available support information into prototypes unavoidably leads to significant spatial information loss. Thus, the matching-based methods are proposed to explore pixel-to-pixel dense connections between support and query images, which can be achieved through graph attention mechanisms~\cite{zhang2019pyramid}, center-pivot 4D convolutions~\cite{min2021hypercorrelation}, and cycle-consistent transformer~\cite{zhang2021few}. The proposed PGMA-Net also belongs to matching-based method, but with a more class-agnostic prior-to-mask mapping.

\subsection{CLIP in FSS task}
Due to the exceptional efficacy in integrating visual and textual features in the CLIP~\cite{radford2021learning} embedding space, there have been attempts to utilize the CLIP prior for few-shot classification tasks~\cite{zhang2022tip,hu2022pushing}. However, CLIP's image-level training leads a discrepancy with dense tasks, rendering it a rising issue to extend in pixel-level tasks~\cite{li2022languagedriven,liu2023delving}. The complexity of this problem is even heightened when taking into account the interactions between the support and query in the FSS task. 

An attempt to incorporate CLIP prior in FSS is CLIPSeg~\cite{luddecke2022image}, involving a simple interaction among support feature, query feature, and textual feature via mix-up operation and FiLM layer~\cite{dumoulin2018feature}. But the generalization is hindered due to the use of such feature-level interaction. Similarly, when applying CLIP prior to FSS task without support mask in IMR-HSNet~\cite{wang2023iterative}, the improper usage of feature-level interaction also leads to limited performance. Compared to these methods, our work aims to better utilize CLIP in FSS task in a more class-agnostic manner, while still having enough flexibility to allow a single set of parameters to perform additional tasks, e.g., zero-shot segmentation, co-segmentation, box-level segmentation, cross-domain FSS, etc.

\section{Methodology}
\subsection{Problem Setup}

Few-shot segmentation devotes to segment novel classes with only a limited number of labeled images. To achieve this, the model is trained on the base categories $\boldsymbol{C}_{b}$ of base dataset $\mathcal{D}_{b}$, and must possess the ability to provide reliable inference on the novel categories $\boldsymbol{C}_{n}$ of novel dataset $\mathcal{D}_{n}$. Noted that the set of base categories $\boldsymbol{C}_{b}$ and novel categories $\boldsymbol{C}_{n}$ are mutually exclusive, i.e., $\boldsymbol{C}_{b} \cap \boldsymbol{C}_{n}=\emptyset$. 

Following previous works~\cite{tian2020prior,min2021hypercorrelation,shi2022dense}, the episodic sampling process is employed in both training and testing stages. Each episode task $\mathcal{T}$ consists of a support set $\mathcal{S}$ and a query set $\mathcal{Q}$, e.g., $\mathcal{T}=(\mathcal{S}, \mathcal{Q})$. The most widely used formulation of an episodic task $\mathcal{T}$ is the $N$-way $K$-shot manner. This entails sampling $N$ classes from the corresponding dataset, with only $K$ labeled images available per class (typically 1 or 5). The few available labeled data are called support set $\mathcal{S}=\left\{\left(I_{s}, M_{s}\right)\right\}_{i=1}^{N \times K}$ and the data waiting for segmenting are called query set $\mathcal{Q}=\left\{\left(I_{q}, M_{q}\right)\right\}_{i=1}^{N}$, where $I_{s} \in \mathbb{R}^{H_{s} \times W_{s} \times 3}$,$I_{q} \in \mathbb{R}^{H_{q} \times W_{q} \times 3}$,$M_{s} \in \mathbb{R}^{H_{s} \times W_{s}}$ and $M_{q} \in \mathbb{R}^{H_{q} \times W_{q}}$ represent the support image, query image, ground-truth mask of support and query respectively.

\begin{figure*}[t!]
  \centering
  \includegraphics[width=1.0 \linewidth]{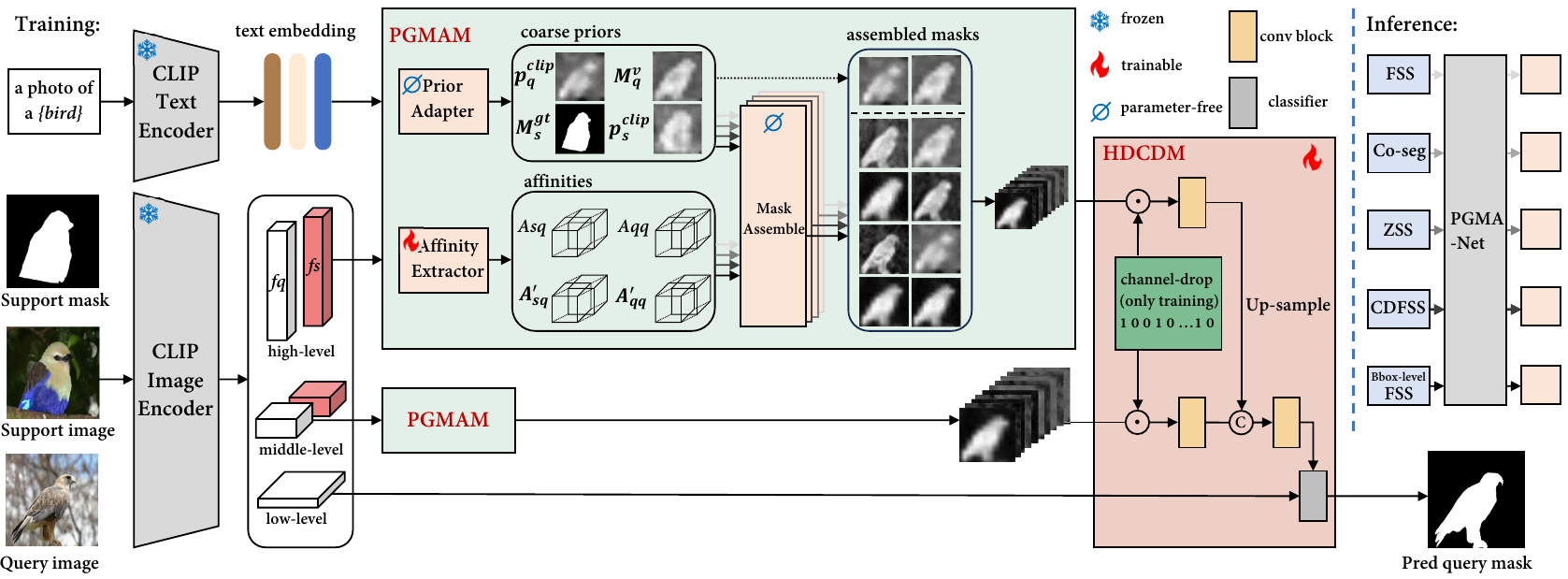}
  \caption{The pipeline of our proposed PGMA-Net: it first extracts class-agnostic priors via Prior Adapter, and affinities via Affinity Extractor. Then performs Mask Assembly via 10 diverse interactions in a unified manner. Finally, the HDCDM is proposed to decode the multiscale assembled masks into prediction mask. Due to the unified way of assembling the coarse prior by affinity, as well as the introduction of channel-drop mechanism, the PGMA-Net trained for FSS is capable of addressing additional tasks with one suit of model weights during inference.}
  \label{fig:framework}
\end{figure*}

\subsection{Prior Guided Mask Assemble Network}

\subsubsection{Overview}
\label{subsubsection:Overview}

As shown in Figure~\ref{fig:framework}, the core of PGMA-Net is a class-agnostic and unified process: assembling the prior by affinity, where the prior in the form of probability map is extracted via Prior Adapter (Section~\ref{subsubsection:PA}), and the affinity is obtained by Affinity Extractor ((Section~\ref{subsubsection:AE})). Then the assembly is executed between coarse priors (from support or query) and affinities (inter- and intra-image, training-free, and high-order) in a unified form, detailed in Section~\ref{subsubsection:assemble}, and decode into the prediction mask via HDCDM (Section~\ref{subsubsection:HDCDM}). Due to the unified way of assembling the coarse prior by affinity, as well as the introduction of channel-drop mechanism, the PGMA-Net trained for FSS is capable of addressing additional tasks with one suit of model weights during inference.

\subsubsection{Prior Adapter}
\label{subsubsection:PA}

\textbf{$\bullet$}\textbf{Textual Prior Adapter}. A straightforward way~\cite{luddecke2022image} to utilize CLIP~\cite{radford2021learning} entails extracting semantically enriched visual features for downstream decoder. However, it is contradictory to our suggested underlying class-agnostic principle. Thus, we opt to first transform them into class-agnostic textual priors without requiring any additional training, and extend CLIP's ability in integration visual and textual features from image-level to spatial positioning. Specifically, prompt engineering is utilized to obtain the text embedding $f_{text} \in \mathbb{R}^{1 \times d}$ for each category, where $d$ is the dimension of the embedding space. Meanwhile, the visual feature $f_{visual} \in \mathbb{R}^{h \times w \times d}$ is obtained by feeding the \textit{value} of patch embedding to the last projection layer of the CLIP model, without the usage of pooling layer. The cosine similarity is calculated between $f_{visual}$ and $f_{text}$, followed by a min-max normalization to highlight the area of interest. The class-agnostic textual prior in terms of probability map is obtained as:
\begin{equation}
  p_{clip} = \operatorname{COS}\left(f_{visual}, f_{text}\right)
\end{equation}
\begin{equation}
  p_{clip}^{'} = \frac{p_{clip}-min(p_{clip})}{max(p_{clip})-min(p_{clip})+\epsilon} 
\end{equation}
where $\epsilon$ is set to 1$e$-8 to avoid division by zero. For brevity, we adopt $p_{clip}$ to denote the normalized visual-textual prior $p_{clip}^{'}$. The resultant CLIP prior is generated for both support and query images with up-sampling to image size, termed as $p_{s}^{clip} \in \mathbb{R}^{H_{s} \times W_{s}} $ and $p_{q}^{clip} \in \mathbb{R}^{H_{q} \times W_{q}}$, respectively.

\textbf{$\bullet$}\textbf{Visual Prior Adapter}. The basic visual prior that needs to undergo assembly is the support-guided visual prior $M_{q}^{v} \in \mathbb{R}^{h_{q} \times w_{q}}$. To obtain $M_{q}^{v}$, the maximum value of each query axis from the cross-affinity $A_{sq}$ is selected, i.e., $M_{q}^{v} = \operatorname{max}_{h_q,w_q} A_{sq}$, followed with a min-max normalization operation to highlight the intended area, where $A_{sq}$ is detailed in Equation~\ref{Asq}. Besides, the annotation of support image can also serve as prior for aggregation. Concretely, we can leverage down-sampled ground-truth mask of support $M_{s}^{gt} \in \mathbb{R}^{h_{s} \times w_{s}}$ and clip prior of support $p_{s}^{clip} \in \mathbb{R}^{h_{s} \times w_{s}}$ as prior.

Taken together, these four priors ($p_{q}^{clip},M_{q}^{v},M_{s}^{gt}, p_{s}^{clip}$) form the basic components for further assembly.

\subsubsection{Affinity Extractor}
\label{subsubsection:AE} 
The cross-affinity between support and query images is the first criterion for assembling. For layer $l$ of the pre-trained backbone, support feature $f_{s}^{l} \in \mathbb{R}^{h_{s}^{l} \times w_{s}^{l} \times d}$, query feature $f_{q}^{l} \in \mathbb{R}^{h_{q}^{l} \times w_{q}^{l} \times d}$ and the down-sampled support mask $M_{s}^{l} \in \mathbb{R}^{h_{s}^{l} \times w_{s}^{l}}$ are obtained. For brevity, the layer index $l$ is omitted. Then a transform is first applied as:
\begin{equation}
  \hat{f}_{s}  = \phi \left({f}_{s} \odot M_{s}\right), 
  \hat{f}_{q}  = \phi \left({f}_{q}\right)
\end{equation}
where $\phi$ indicates the reshape operation: $\mathbb{R}^{h \times w \times d} \rightarrow \mathbb{R}^{h w \times d}$, and $\odot$ denotes Hadamard product to discard the irrelevant features. Then the $4d$ cross-affinity $A_{sq} \in \mathbb{R}^{h_{s} \times w_{s} \times h_{q}\times w_{q}}$ is calculated as:
\begin{equation}
  A_{sq} = \frac{{\hat{f}_{s}}^{T} \hat{f}_{q}}{\left\|\hat{f}_{s}\right\|\left\|\hat{f}_{q}\right\|}
  \label{Asq} 
\end{equation}
where $T$ indicates feature transpose. Previous methods ~\cite{min2021hypercorrelation,shi2022dense} rely solely on this cross-affinity $A_{sq}$ for segmentation, we contend that it's insufficient as this criterion ignores the self-affinity of both support and query images, which represent intrinsic structural information that is indispensable to understand an image. So the self-affinity of support $A_{ss} \in \mathbb{R}^{h_{s}\times w_{s} \times h_{s}\times w_{s}}$ and that of query $A_{qq} \in \mathbb{R}^{h_{q}\times w_{q} \times h_{q}\times w_{q}}$ are formulated identically as:
\begin{equation}
  A_{ss} = \frac{{\hat{f}_{s}}^{T} \hat{f}_{s}}{\left\|\hat{f}_{s}\right\|\left\|\hat{f}_{s}\right\|} , A_{qq} = \frac{{\hat{f}_{q}}^{T} \hat{f}_{q}}{\left\|\hat{f}_{q}\right\|\left\|\hat{f}_{q}\right\|} 
\end{equation}

\begin{figure}[t]
  \centering
  \includegraphics[width=1.0 \linewidth]{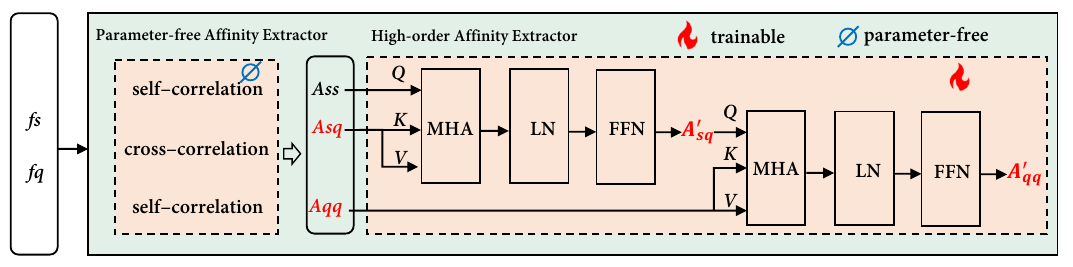}
  \caption{The structure of our proposed affinity extractor, including parameter-free affinity extractor and high-order affinity extractor.}
  \label{fig:ae}
\end{figure}

The above three affinities are generated by parameter-free process, implying a simple yet effective set of criteria. However, this also limits their capacity for accommodating large variations among text, support and query~\cite{ZhangWWG22}. Therefore, we propose a high-order affinity extractor to enlarge its capacity, implemented by multi-head cross-attention mechanism~\cite{vaswani2017attention}. The high-order version of cross-affinity $A_{sq}^{'}$ is calculated as:
\begin{equation}
  A_{sq}^{'} = \operatorname{FFN}\left(\operatorname{LN}\left(\operatorname{MHA}\left(A_{ss}, A_{sq},A_{sq}\right)\right)\right)
\end{equation}
where the affinity $A_{ss}$ is considered as \textit{QUERY}\footnote[1]{The \textit{QUERY} in attention mechanism, is different from the query image in few-shot segmentation task, we use uppercase italics "\textit{QUERY}" and regular text "query" to distinguish them.} sequence with the length of $h_s\times w_s$, and in dim of $h_s\times w_s$. And the affinity $A_{sq}$ is considered as the \textit{KEY} and \textit{VALUE} sequence with the length of $h_q\times w_q$, and in dim of $h_s\times w_s$. The $\operatorname{FFN}$, $\operatorname{LN}$ and $\operatorname{MHA}$ denote the feed-forward network, layer normalization and multi-head attention respectively. Similarly, the high-order version of self-affinity $A_{qq}^{'}$ is calculated by considering $A_{sq}^{'}$ as \textit{QUERY}, $A_{qq}$ as \textit{KEY} and \textit{VALUE}:
\begin{equation}
  A_{qq}^{'} = \operatorname{FFN}\left(\operatorname{LN}\left(\operatorname{MHA}\left(A_{sq}^{'}, A_{qq},A_{qq}\right)\right)\right)
\end{equation}
The high-order affinity extractor differs from~\cite{ZhangWWG22} in training-free input affinity, extra outputs (high-order self-affinity) and subsequent class-agnostic usage. In total, four affinities are obtained: $(A_{sq}, A_{qq},A_{sq}^{'},A_{qq}^{'})$.

\subsubsection{Prior-Guided Mask Assemble Module}
\label{subsubsection:assemble}

Once priors are obtained, a straightforward way to explore such priors is to multiply them with image features at pixel-level via Hadamard product, yielding refined semantically rich features corresponding to its category by pooling operation. However, such a naive operation does not adhere to our suggested class-agnostic principle, and fails to fully exploit the possible interactions (visual-textual, intra- or inter-image), along with issue of information loss due to the pooling operation. Thus, we propose a new unified Prior-Guided Mask Assemble Module (PGMAM), achieving the class-agnostic prior-to-mask mapping by assembling diverse priors according to corresponding affinities at pixel-level.

\begin{figure}[h]
  \centering
  \includegraphics[width=1.0 \linewidth]{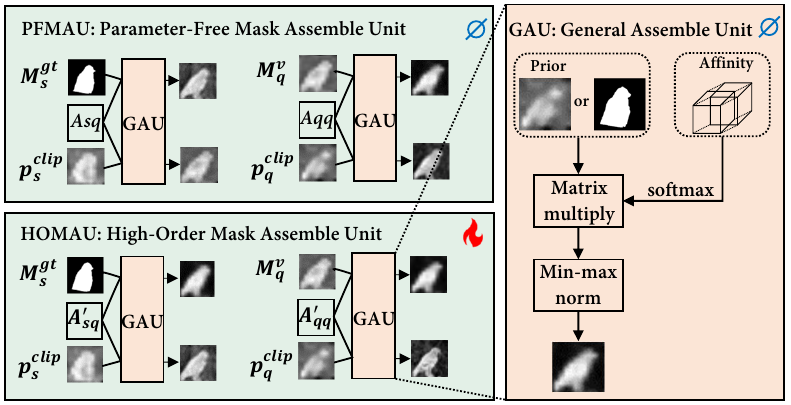}
  \caption{The pipeline of PGMAM (consisting multiple GAUs): it achieves a class-agnostic assembly of diverse priors and corresponding affinities at pixel-level in a unified way. Regardless of whether prior is from support or query, the GAU consistently follows the pipeline of emphasizing the affinity along a specific axis and then integrates it through matrix multiplication and normalization process.}
  \label{fig:assemble}
\end{figure}

As shown in Figure~\ref{fig:assemble}, the GAU follows the pipeline of highlighting the affinity along a particular axis, and then integrating through matrix multiplication and min-max normalization:
\begin{equation}
  p_{refined} = \operatorname{SoftMax}\left(A\right) \cdot p
\end{equation}
This module is unified regardless of whether the input affinity is parameter-free ($A_{sq}$ and $A_{qq}$) or high-order ($A_{sq}^{'}$ and $A_{qq}^{'}$), or whether the prior is from support ($p_{q}^{clip},M_{q}^{v}$) or query ($M_{s}^{gt}, p_{s}^{clip}$). We refer the GAU as parameter-free mask assemble unit (PFMAU) when the affinity is parameter-free, and high-order mask assemble unit (HOMAU) when the affinity is high-order.

\begin{figure}[t]
  \centering
  \includegraphics[width=1.0 \linewidth]{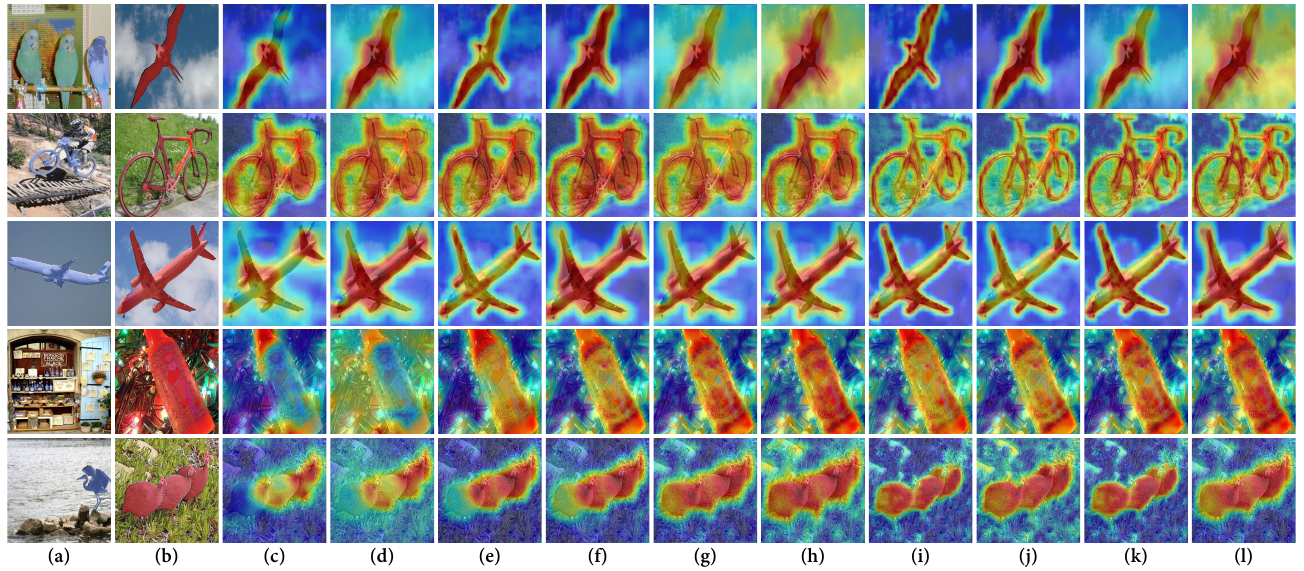}
  \caption{Visualization of heterogeneous interactions among diverse priors and affinities using GAU. Columns a to b: support and query images, c to d:  coarse prior of the query obtained through visual-textual and support-query, e to h: assembled priors by PFMAU, and i to l: assembled priors by HOMAU.}

  \label{fig:diverse_interaction}
\end{figure}

Figure~\ref{fig:diverse_interaction} presents visual representations of pre- and post-assemble states of the clip prior $p_{q}^{clip}$ (column c) and support-guided visual prior $M_{q}^{v}$ (column d) using GAU. The proposed GAU has the potential to assemble the coarse priors into a refined prior by utilizing the 8 types of diverse interactions. The 8 assembled priors are: the parameter-free self-affinity guided prior $A_{qq} \cdot p_{q}^{clip}$ and $A_{qq} \cdot M_{q}^{v}$, the high-order self-affinity guided prior $A_{qq}^{'} \cdot p_{q}^{clip}$ and $A_{qq}^{'} \cdot M_{q}^{v}$, the parameter-free cross-affinity guided prior $A_{sq}^{T} \cdot M_{s}^{gt}$ and $A_{sq}^{T} \cdot p_{s}^{clip}$, and the high-order cross-affinity guided prior $A_{sq}^{'T} \cdot M_{s}^{gt}$ and $A_{sq}^{'T} \cdot p_{s}^{clip}$. These total 10 priors, each containing distinct information, are concatenated and subsequently fed into a hierarchical decoder for decoding into final segmentation results:
\begin{equation}
  P_{refine} = \operatorname{CONCAT}(p_{q}^{clip},M_{q}^{v},\dots,A_{sq}^{'T} \cdot p_{s}^{clip})
\end{equation}

\subsubsection{Hierarchical decoder with channel-drop mechanism}
\label{subsubsection:HDCDM}
By employing the proposed PGMAM on multiscale features from various layers of a pre-trained backbone, several sets of assembled masks $(\dots,P_{refine}^{l-1},P_{refine}^{l})$ are generated, with high-level and low-level $P_{refine}$ containing semantic and fine-grained information, respectively. To maximize inter-scale information utilization, a hierarchical decoder (HDCDM) is designed to decode these assembled masks from varying scales into segmentation outputs. Specifically, as shown in Figure~\ref{fig:framework}, this process involves initial feature transformation of the assembled masks at each scale using a convolution block, and if necessary, up-sampling before concatenation with higher-resolution features. 

Considering that each group of $P_{refine}$ constitutes assembled masks from different prior information sources with varying learning difficulties, to avoid the network overfitting to some simple priors and maximize the utilization of all assembly masks, we propose a channel-drop mechanism during training:
\begin{equation}
  P^{l'}_{refine} = \operatorname{Channel-Drop}_{\pi}(P^{l}_{refine})
\end{equation}
where the channel-drop probability $\pi$ is a random vector of $0$ and $1$, with the same length as channel number in $P^{l}_{refine}$. Such channel-drop mechanism seamlessly incorporates various scenarios such as fully-supervised FSS, FSS without support mask, and zero-shot segmentation tasks into a unified framework.

The training loss $\mathcal{L}$ was a weighted sum of cross-entropy loss $\mathcal{L}_{ce}$ and dice loss $\mathcal{L}_{dice}$:
\begin{equation}
  \mathcal{L} = \lambda \mathcal{L}_{dice} + (1-\lambda) \mathcal{L}_{ce}
\end{equation}
\begin{equation}
  \mathcal{L}_{dice}= 1 - \frac{2\times\sum_{i=1}^{H}\sum_{j=1}^{W}y_{i,j}\times\hat{y}_{i,j}}{\sum_{i=1}^{H}\sum_{j=1}^{W}y_{{i,j}}^2 + \sum_{i=1}^{H}\sum_{j=1}^{W}\hat{y}_{i,j}^2}
\end{equation}
\begin{equation}
  \mathcal{L}_{ce} = -\frac{1}{H\times W}\sum_{i=1}^{H}\sum_{j=1}^{W}(y_{i,j}\log( \hat{y}_{i,j}) + (1-y_{i,j})\log(1-\hat{y}_{i,j}))
\end{equation}
where the weight $\lambda$ is set to 0.5, $y_{i,j}$ is the ground-truth label, $\hat{y}_{i,j}$ is the predicted label.

\begin{table*}
  \centering
  \caption{Comparison of the proposed PGMA-Net with the current SOTA on $\text{PASCAL-}5^i$ ~\cite{shaban2017one}. Best results are in bold.}
  \label{tab:pascal_sota}
      \scalebox{0.85}{
      \begin{tabular}{cclc|cccccc|cccccc}
        \toprule
        \multirow{2}{*}{\shortstack{\textbf{Pretrain}}} & \multirow{2}{*}{\shortstack{\textbf{Backbone}}} & \multirow{2}{*}{\textbf{Method}} & \multirow{2}{*}{\textbf{Publication}} & \multicolumn{6}{c}{\textbf{1-shot}} & \multicolumn{6}{c}{\textbf{5-shot}}  \\ 
        
        & & & & $5^{0}$ & $5^{1}$ & $5^{2}$ & $5^{3}$ & \textbf{mIoU} & \textbf{FB-IoU} & $\mathbf{5^{0}}$ & $\mathbf{5^{1}}$ & $\mathbf{5^{2}}$ & $\mathbf{5^{3}}$ & \textbf{mIoU} & \textbf{FB-IoU}  \\
        \midrule

        \multirow{22}{*}{IN1K} & \multirow{12}{*}{RN50}  & PPNet~\cite{liu2020part} & ECCV'20     & 48.6 & 60.6 & 55.7 & 46.5 & 52.8 & 69.2 & 58.9 & 68.3 & 66.8 & 58.0 & 63.0 & 75.8 \\  
        
        && PFENet~\cite{tian2020prior} & TPAMI'20  & 61.7 & 69.5 & 55.4 & 56.3 & 60.8 & 73.3 & 63.1 & 70.7 & 55.8 & 57.9 & 61.9 & 73.9  \\ 
        
        && RePRI~\cite{boudiaf2021few} & CVPR'21 & 59.8 & 68.3 & 62.1 & 48.5 & 59.7 & - & 64.6 & 71.4 & 71.1 & 59.3 & 66.6 & -  \\
        
        && HSNet~\cite{min2021hypercorrelation} & ICCV'21   & 64.3 & 70.7 & 60.3 & 60.5 & 64.0 & 76.7 & 70.3 & 73.2 & 67.4 & 67.1 & 69.5 & 80.6  \\

        && SSP~\cite{fan2022self}  & ECCV'22   & 60.5 & 67.8& 66.4 & 51.0& 61.4& - & 67.5 & 72.3 & 75.2 & 62.1 & 69.3 & - \\

        && DCAMA~\cite{shi2022dense} & ECCV'22  & 67.5 & 72.3 & 59.6 & 59.0 & 64.6 & 75.7 & 70.5 & 73.9 & 63.7 & 65.8 & 68.5 & 79.5  \\  

        && CATrans~\cite{ZhangWWG22} & IJCAI'22  & 67.6 & 73.2 & 61.3 & 63.2 & 66.3 & - & 75.1 & 78.5 & 75.1 & 72.5 & 75.3 & -  \\  

        && DPCN~\cite{liu2022dynamic} & CVPR'22 & 65.7 & 71.6 & 69.1 & 60.6 & 66.7 & 78.0 & 70.0 & 73.2 & 70.9 & 65.5 & 69.9 & 80.7  \\  
        
        && RPMG-FSS~\cite{zhang2023rpmg} & TCSVT'23  & 64.4 & 72.6 & 57.9 & 58.4 & 63.3 & - & 65.3 & 72.8 & 58.4 & 59.8 & 64.1 & -  \\  

        && HPA\cite{cheng2023hpa} & TPAMI'23  & 65.9 & 72.0 & 64.7 & 56.8 & 64.8 & 76.4 & 70.5 & 73.3 & 68.4 & 63.4 & 68.9 & 81.1  \\  

        && FECANet~\cite{liu2023fecanet} & TMM'23  & 69.2 & 72.3 & 62.4 & 65.7 & 67.4 & 78.7 & 72.9 & 74.0 & 65.2 & 67.8 & 70.0 & 80.7  \\

        && ABCNet~\cite{Wang_2023_CVPR} & CVPR'23 &68.8&73.4&62.3&59.5&66.0&76.0&71.7&74.2&65.4&67.0&69.6&80.0 \\  
        \cline{2-16} \\[-2.0ex]

        & \multirow{10}{*}{RN101} & PPNet~\cite{liu2020part} & ECCV'20   & 52.7 & 62.8 & 57.4 & 47.7 & 55.2 & 70.9 & 60.3 & 70.0 & 69.4 & 60.7 & 65.1 & 77.5  \\  
        && DAN~\cite{wang2020few} & ECCV'20   & 54.7 & 68.6 & 57.8 & 51.6 & 58.2 & 71.9 & 57.9 & 69.0 & 60.1 & 54.9 & 60.5 & 72.3  \\ 
        && PFENet~\cite{tian2020prior} & TPAMI'20  & 60.5 & 69.4 & 54.4 & 55.9 & 60.1 & 72.9 & 62.8 & 70.4 & 54.9 & 57.6 & 61.4 & 73.5\\
        && RePRI~\cite{boudiaf2021few} & CVPR'21 & 59.6 & 68.6 & 62.2 & 47.2 & 59.4 & - & 66.2 & 71.4 & 67.0 & 57.7 & 65.6 & -  \\ 
        && HSNet~\cite{min2021hypercorrelation} & ICCV'21    & 67.3 & 72.3 & 62.0 & 63.1 & 66.2 & 77.6 & 71.8 & 74.4 & 67.0 & 68.3 & 70.4 & 80.6  \\
        && SSP~\cite{fan2022self}  & ECCV'22   & 63.7 & 70.1& 66.7 & 55.4& 64.0& - & 70.3 & 76.3 & 77.8 & 65.5 & 72.5 & -  \\
        && DCAMA~\cite{shi2022dense} & ECCV'22  & 65.4 & 71.4 & 63.2 & 58.3 & 64.6 & 77.6 & 70.7 & 73.7 & 66.8 & 61.9 & 68.3 & 80.8  \\  
        && RPMG-FSS~\cite{zhang2023rpmg} & TCSVT'23  & 63.0 & 73.3 & 56.8 & 57.2 & 62.6 & - & 67.1 & 73.3 & 59.8 & 62.7 & 65.7 & -  \\  
        && HPA~\cite{cheng2023hpa} & TPAMI'23  & 66.4 & 72.7 & 64.1 & 59.4 & 65.6 & 76.6 & 68.0 & 74.6 & 65.9 & 67.1 & 68.9 &80.4 \\
        && ABCNet~\cite{Wang_2023_CVPR} & CVPR'23&65.3&72.9&65.0&59.3&65.6&78.5&71.4&75.0&68.2&63.1&69.4&80.8\\ 
        
        \midrule 

        \multirow{5}{*}{CLIP} &CLIP-ViT-B/16&  CLIPSeg(PC+)~\cite{luddecke2022image} & CVPR'22 & - & - & - & - & 59.5 & - & - & - & - & - & - & - \\
        &CLIP-ViT-B/16&  CLIPSeg(PC)~\cite{luddecke2022image} & CVPR'22 & - & - & - & - & 52.3 & - & - & - & - & - & - & - \\ 
        
        &CLIP-ViT-B/16& PGMA-Net (ours)  & - & 74.0 & 81.9 & 66.8 & 73.7 & 74.1 & 82.1 & 74.5 & 82.2 & 67.2 & 74.4 & 74.6 & 82.5 \\
        
        &CLIP-RN50& PGMA-Net (ours) & - & 73.4 & 80.8 & 70.5 & 71.7 & 74.1 & 83.5 & 74.0 & 81.5 & 71.9 & 73.3 & 75.2 & 84.2 \\
        
        &CLIP-RN101& PGMA-Net (ours)  &-&  \textbf{76.8} & \textbf{82.3} & \textbf{75.7} & \textbf{75.7} & \textbf{77.6} & \textbf{86.2} & \textbf{77.7} & \textbf{82.7} & \textbf{76.9} & \textbf{77.0} & \textbf{78.6} & \textbf{86.9} \\

        \bottomrule

      \end{tabular}
      }
  \hfill

\end{table*}

\section{Experiments}

\subsection{Datasets Setup and Evaluation Metrics}
Following the setup of~\cite{tian2020prior,min2021hypercorrelation}, we evaluated PGMA-Net on two benchmarks:  $\text{PASCAL-}5^i$~\cite{shaban2017one} and $\text{COCO-}20^i$~\cite{nguyen2019feature}.  $\text{PASCAL-}5^i$ is created by combining the PASCAL VOC 2012 ~\cite{everingham2010pascal} and SBD datasets~\cite{hariharan2011semantic}, consisting of 20 classes evenly distributed in 4 folds with 5 classes per fold. $\text{COCO-}20^i$ is a larger and more challenging benchmark created from the COCO~\cite{lin2014microsoft}, consisting of 80 classes evenly divided into 4 folds. For all experiments, we selected a pre-defined set of three folds for training, while reserving the remaining fold exclusively for evaluation purposes.

The evaluation metrics used in the experiments were the mean intersection over union ($\mathrm{mIoU}=\frac{1}{C} \sum_{c=1}^{C} \operatorname{IoU}{c}$) and foreground-background IoU ($\text{FB-IoU}=\frac{1}{2}(\mathrm{IoU}{\mathrm{F}}+\mathrm{IoU}_{\mathrm{B}})$). We assessed the performance of the model by conducting experiments for each fold and reporting the mean mIoU and FB-IoU across all folds.

\subsection{Implemetation Details}
All experiments were implemented via PyTorch framework on a single NVIDIA GeForce RTX 3090 GPU. The support and query images utilized an equivalent input resolution of $384 \times 384$ pixels. The data augmentation techniques included random scales, rotations and flips. We utilized the CLIP-RN50, CLIP-RN101, CLIP-ViT-B/16, IN1K-RN50 as our backbone. As for the optimization and scheduling, we trained our model with episodic training scheme for 100 and 200 epochs on $\text{PASCAL-}5^i$ and $\text{COCO-}20^i$ datasets, respectively. AdamW served as our optimizer, with a learning rate of $0.001$. We determined batch sizes of 6 and 4 for $\text{PASCAL-}5^i$ and $\text{COCO-}20^i$ datasets, respectively.

\subsection{Comparison with State-of-the-Art Methods}

\textbf{FSS task.}
Experiments were conducted on $\text{PASCAL-}5^i$~\cite{shaban2017one} and $\text{COCO-}20^i$~\cite{nguyen2019feature} datasets. Table~\ref{tab:pascal_sota} presents the mIoU and FB-IoU results obtained on $\text{PASCAL-}5^i$~\cite{shaban2017one} using 1-shot and 5-shot settings. With CLIP-RN50 backbone, we achieved mIoU of 74.1.  When using the same CLIP-ViT-B/16 backbone, PGMA-Net achieved mIoU of 74.1 in 1-shot scenario, outperforming CLIPSeg~\cite{luddecke2022image} by a large margin (14.6 mIoU increase). Our model is the first to mitigate the performance gap between previous CLIP-based and traditional FSS, and surpass both subfields by a large margin [our 74.1 v.s. CLIPSeg~\cite{luddecke2022image} 59.5 v.s. FECANet~\cite{liu2023fecanet} 67.4].

\begin{figure}[h]
  \centering
  \includegraphics[width=1.0 \linewidth]{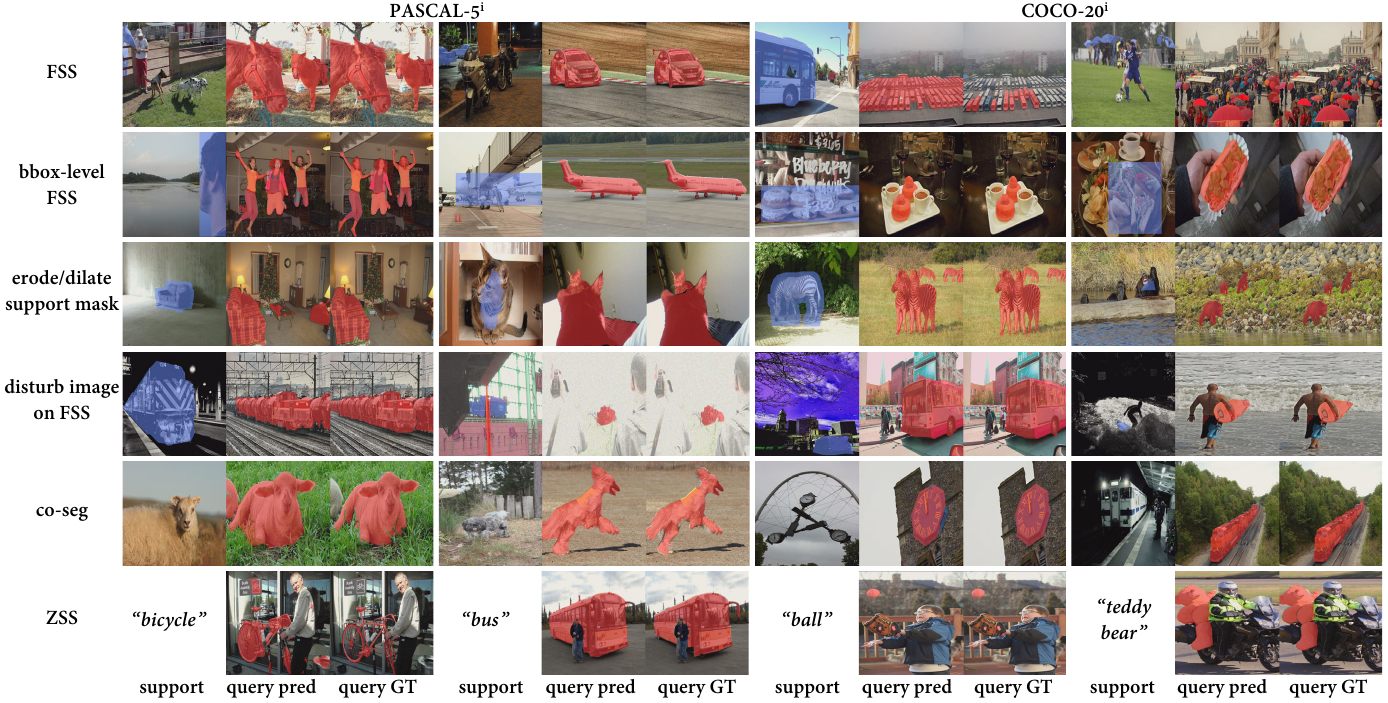}
  \caption{Visualizations of the proposed PGMA-Net on different tasks. (Zoom for a better view). The proposed PGMA-Net trained for few-shot segmentation task has the capability to perform any-shot segmentation tasks via a single set of parameters. Rows 1-6 are FSS, bbox-level FSS, FSS with eroded/dilated support mask, FSS with noise and distortion in images, co-segmentation and zero-shot segmentation tasks, respectively.}
  \label{fig:vis}
\end{figure}

Table~\ref{tab:coco_sota} displays significant and consistent improvements on the $\text{COCO-}20^i$ dataset ~\cite{nguyen2019feature}. With CLIP-RN101 backbone, our PGMA-Net achieved mIoU values of 59.4 for 1-shot scenario. Our PGMA-Net outperforms previous CLIP-based CLIPSeg~\cite{luddecke2022image} (mIoU of 33.3) and IN-1K based FSS (HPA~\cite{cheng2023hpa}, mIoU of 45.8) by a large margin. The top row of visualizations presented in Figure~\ref{fig:vis} showcases several successful instances. 

\textbf{Cross-domain FSS task.}
To illustrate the effectiveness of the proposed PGMA-Net in handling cross-domain issues, we conducted experiments on the $\text{COCO-}20^i$~\cite{nguyen2019feature} to $\text{PASCAL-}5^i$~\cite{shaban2017one} task, following the experimental setup of CDFSS~\cite{wang2022remember}. Table~\ref{tab:cross-domain-bbox-image} indicates that, despite not being designed specifically for cross-domain problem, PGMA-Net achieved a mIoU increase of 6.8, compared to the current SOTA method CDFSS (72.4 \text{v.s.} 65.6). The robustness can be attributed that both the diverse priors and affinities exhibit diminished disparity across varied datasets.

\textbf{Bounding-box level FSS task.}
Annotating a few support images at the pixel level is also  time-consuming. One feasible solution is to use weaker annotations as segmentation clues, e.g., bounding-box. To accomplish this, we filled in the provided bounding-box to generate pseudo mask~\cite{wang2019panet,Zhang2019canet}, which can be directly tested by the previously trained model without requiring retraining. Table~\ref{tab:cross-domain-bbox-image} and Figure~\ref{fig:vis} demonstrate the effectiveness of PGMA-Net, which shows a significant increase in absolute mIoU by 12.0 compared to DCAMA~\cite{shi2022dense} (73.2 \text{v.s.} 61.2).

\begin{table*}[t]
  \centering
  \caption{Comparison of the proposed PGMA-Net with the current SOTA on $\text{COCO-}20^i$ dataset~\cite{nguyen2019feature}. Best results are in bold.}
  \label{tab:coco_sota}
      \scalebox{0.85}{
      \begin{tabular}{cclc|cccccc|cccccc}
        \toprule
        \multirow{2}{*}{\shortstack{\textbf{Pretrain}}} & \multirow{2}{*}{\shortstack{\textbf{Backbone}}} & \multirow{2}{*}{\textbf{Method}} & \multirow{2}{*}{\textbf{Publication}} & \multicolumn{6}{c}{\textbf{1-shot}} & \multicolumn{6}{c}{\textbf{5-shot}}  \\ 
        & & & & $\mathbf{20^{0}}$ & $\mathbf{20^{1}}$ & $\mathbf{20^{2}}$ & $\mathbf{20^{3}}$ & \textbf{mIoU} & \textbf{FB-IoU} & $\mathbf{20^{0}}$ & $\mathbf{20^{1}}$ & $\mathbf{20^{2}}$ & $\mathbf{20^{3}}$ & \textbf{mIoU} & \textbf{FB-IoU}  \\
        
        \midrule
        \multirow{17}{*}{IN1K} & \multirow{12}{*}{RN50}
        & PPNet~\cite{liu2020part} & ECCV'20    & 28.1 & 30.8 & 29.5 & 27.7 & 29.0 & - & 39.0 & 40.8 & 37.1 & 37.3 & 38.5 & -   \\  
        && PFENet~\cite{tian2020prior} & TPAMI'20   & 36.5 & 38.6 & {34.5} & {33.8} & {35.8} & - & 36.5 & 43.3 & 37.8 & 38.4 & 39.0   \\ 
        && RePRI~\cite{boudiaf2021few} & CVPR'21 & 32.0 & 38.7 &  32.7 & 33.1 & 34.1 & - & 39.3 & 45.4 & 39.7 & 41.8 & 41.6 & - \\
        && HSNet~\cite{min2021hypercorrelation} & ICCV'21    & 36.3 & 43.1 & 38.7 & 38.7 & 39.2 & 68.2 & 43.3 & 51.3 & 48.2 & 45.0 & 46.9 &70.7 \\
        && NTRENet~\cite{liu2022learning} & CVPR'22    & 36.8 & 42.6 & 39.9 & 37.9 & 39.3 & 68.5 & 38.2 & 44.1 & 40.4 & 38.4 & 40.3 & 69.2  \\  
        && SSP~\cite{fan2022self}  & ECCV'22   & 35.5 & 39.6& 37.9 & 36.7& 37.4& - & 40.6 & 47.0 & 45.1 & 43.9 & 44.1 & -  \\
        && DCAMA~\cite{shi2022dense} & ECCV'22  & 41.9 & 45.1 & 44.4 & 41.7 & 43.3 & 69.5 & 45.9 & 50.5 & 50.7 & 46.0 & 48.3 & 71.7  \\  
        && DPCN~\cite{liu2022dynamic} & CVPR'22 & 42.0 & 47.0 & 43.2 & 39.7 & 43.0 & 63.2 & 46.0 & 54.9 & 50.8 & 47.4 & 49.8 & 67.4  \\  
        && RPMG-FSS~\cite{zhang2023rpmg} & TCSVT'23  & 38.3 & 41.4 & 39.6 & 35.9 & 38.8 & - & - & - & - & -  \\  
        && HPA~\cite{cheng2023hpa} & TPAMI'23  & 40.3 & 46.6 & 44.1 & 42.7 & 43.4 & 68.2 & 45.5 & 55.4 & 48.9 & 50.2 & 50.0 & 71.2  \\  
        && FECANet~\cite{liu2023fecanet} & TMM'23  & 38.5 & 44.6 & 42.6 & 40.7 & 41.6 & 69.6 & 44.6 & 51.5 & 48.4 & 45.8 & 47.6 & 71.1   \\  
        &&ABCNet~\cite{Wang_2023_CVPR} & CVPR'23&42.3&46.2&46.0&42.0&44.1&69.9&45.5&51.7&52.6&46.4&49.1&72.7\\
        \cline{2-16} \\[-2.0ex]
        
        &\multirow{5}{*}{RN101} & PFENet~\cite{tian2020prior} & TPAMI'20  & {36.8} & {41.8} & {38.7} & {36.7} & {38.5} & {63.0} & {40.4} & {46.8} & {43.2} & {40.5} & {42.7} & {65.8} \\
        && HSNet~\cite{min2021hypercorrelation} & ICCV'21   & {37.2} & {44.1} & {42.4} & {41.3} & {41.2} & {69.1} & {45.9} & {53.0} & {51.8} & {47.1} & {49.5} & {72.4} \\
        && SSP~\cite{fan2022self}  & ECCV'22   & 39.1 & 45.1& 42.7 & 41.2& 42.0& - & 47.4 & 54.5 & 50.4 & 49.6 & 50.2 & -  \\
        && DCAMA~\cite{shi2022dense} & ECCV'22  & 41.5 & 46.2 & 45.2 & 41.3 & 43.5 & 69.9 & 48.0 & 58.0 & 54.3 & 47.1 & 51.9 & 73.3  \\  
        && HPA\cite{cheng2023hpa} & TPAMI'23  & 43.1 & 50.0 & 44.8 & 45.2 & 45.8 & 68.4 & 49.2 & 57.8 & 52.0 & 50.6 & 52.4 & 74.0  \\  
        \midrule 
     
        \multirow{5}{*}{CLIP} &CLIP-ViT-B/16&  CLIPSeg(COCO)~\cite{luddecke2022image} & CVPR'22 & - & - & - & - & 33.2 & - & - & - & - & - & - & - \\
        &CLIP-ViT-B/16&  CLIPSeg(COCO+N)~\cite{luddecke2022image} & CVPR'22 & - & - & - & - & 33.3 & - & - & - & - & - & - & - \\ 
              
        &CLIP-RN50& PGMA-Net (ours) &- &  49.9 & 56.7 & 55.8 & 54.7 & 54.3 & 75.8 & 49.5 & 61.7 & 59.1 & 57.9 & 57.1 & 76.7 \\
        
        &CLIP-RN101& PGMA-Net (ours)  &-&  \textbf{55.2} & \textbf{62.7} & \textbf{60.3} & \textbf{59.4} & \textbf{59.4} & \textbf{78.5} & \textbf{55.9} & \textbf{65.9} & \textbf{63.4} & \textbf{61.9} & \textbf{61.8} & \textbf{79.4} \\

        \bottomrule
      \end{tabular}
      }

  \hfill
\end{table*}

\begin{table*}[h]
  \centering
  \caption{Comparison of zero-shot segmentation task on $\text{PASCAL-}5^i$~\cite{shaban2017one}.}
  \label{tab:zss}%
  \scalebox{0.8}{ 
        \begin{tabular}{cccc|cccccc}
        \toprule
        \textbf{Method} &\textbf{Backbone} &\textbf{Using CLIP} & \textbf{Publication} & $\mathbf{5^{0}}$ & $\mathbf{5^{1}}$ & $\mathbf{5^{2}}$ & $\mathbf{5^{3}}$ & $\textbf{mIoU}$ & $\textbf{FB-IoU}$\\
        \midrule
        SPNet~\cite{xian2019semantic} & RN101 &no & CVPR'19 &23.8&17.0&14.1&18.3&18.3&44.3 \\
        ZS3Net~\cite{bucher2019zero} & RN101 &no & NeurIPS'19 &40.8&39.4&39.3&33.6&38.3&57.7 \\
        LSeg~\cite{li2022languagedriven} & RN101 &yes &ICLR'22 &  52.8&53.8&44.4&38.5&47.4&64.1  \\ 
        LSeg~\cite{li2022languagedriven} & ViT-L/16 &yes &ICLR'22  &61.3&63.6&43.1&41.0&52.3&67.0 \\ 
        SAZS~\cite{liu2023delving} & DRN  &yes &CVPR'23  &57.3&60.3&58.4&45.9&55.5&66.4 \\ 
        SAZS~\cite{liu2023delving} & ViT-L &yes &CVPR'23  &62.7&64.3&60.6&50.2&59.4&69.0 \\ 
        \midrule
        PGMA-Net (ours) w/o fine-tuning & RN50 &yes & -& $54.3$ & $67.7$ & $57.5$ & $60.9$ & $60.1$ & $72.1$  \\ 
        PGMA-Net (ours) -retrain for ZSS &  RN50 &yes &-& $\textbf{68.2}$ & $\textbf{78.8}$ & $\textbf{68.8}$ & $\textbf{66.5}$ & $\textbf{70.6}$ & $\textbf{80.0}$  \\ 
       \bottomrule

        \end{tabular}}
\end{table*}

\textbf{Co-segmentation task.}
An even more challenging task is FSS without support mask~\cite{wang2023iterative, siam2020weakly}, where solely the support image serves as guidance clue. This setup exhibits similarity to co-segmentation~\cite{zhu2020adacoseg,li2019group}, but imposes a more significant challenge in terms of generalization for evaluation on novel categories. As evidenced by Table~\ref{tab:cross-domain-bbox-image} and Figure~\ref{fig:vis}, upon being equipped with the same CLIP-RN50 backbone, albeit not specifically designed or trained for this task, PGMA-Net had already surpassed IMR-HSNet~\cite{wang2023iterative} (intended for this task) by a notable margin. (67.3 \text{v.s.} 61.5)

\textbf{Zero-Shot Segmentation Task.}
Besides, due to the integration of the channel-drop mechanism and textual prior, a singular set of parameters trained for FSS exhibits adequate capacity and flexibility to perform ZSS task. Table~\ref{tab:zss} and Figure~\ref{fig:vis} provide evidence that, even when directly applied to ZSS, our proposed PGMA-Net already outperformed the current SAZS~\cite{liu2023delving} (60.1 \text{v.s.} 59.4). Moreover, our method's superiority became unequivocal (70.6 \text{v.s.} 59.4) when it's trained specifically for ZSS.

\begin{figure}[htbp]
  \centering
  \subfloat[]{
  \label{fig:robust_1}
  \includegraphics[width=0.2\textwidth]{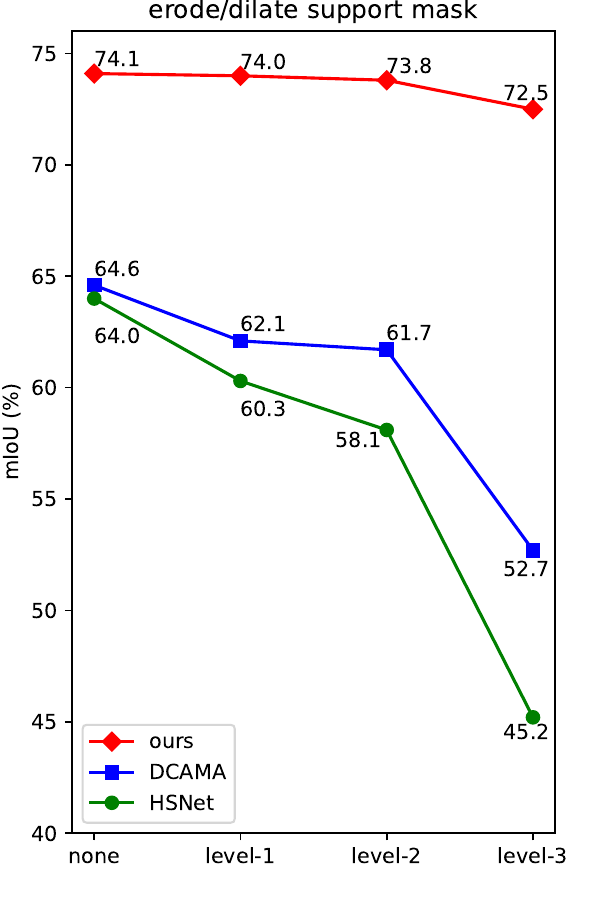}}
  \hspace{0.01\textwidth}
  \subfloat[]{\label{fig:robust_2}
  \includegraphics[width=0.2\textwidth]{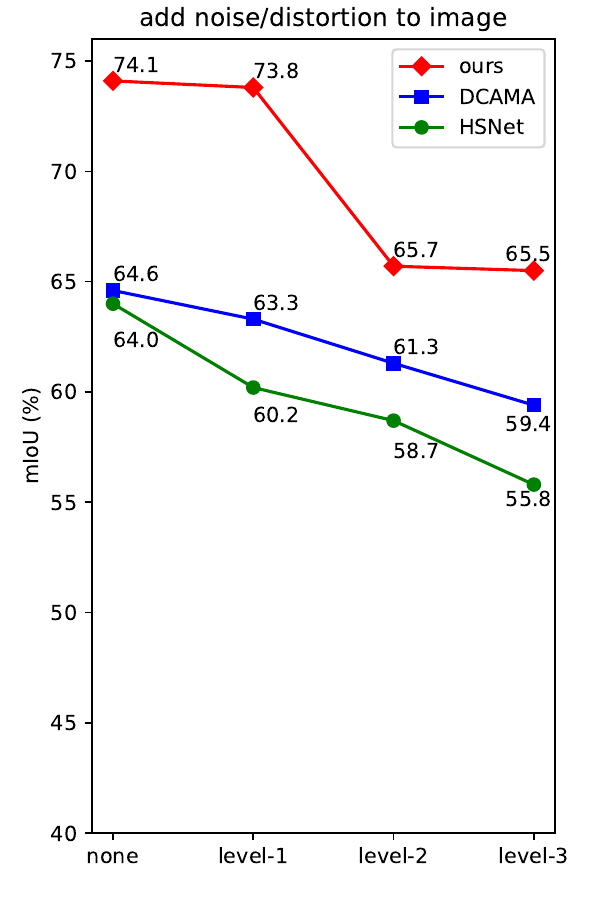}}
  \caption{The robustness of PGMA-Net under (a): three levels of inaccurate support mask with corruption generated by applying random erosion and dilation. (b): three levels of corruption of adding noise and distortion to image.}
  \label{fig:robustness}
\end{figure}

\textbf{Robustness agsinst inaccurate support mask and image quality.}
To demonstrate the robustness of PGMA-Net against inaccurate support mask, random erosion and dilations were applied to the support ground-truth mask, three levels of corruption were implemented. As illustrated in Figure~\ref{fig:robust_1}, PGMA-Net surpassed the DCAMA~\cite{shi2022dense} and HSNet~\cite{min2021hypercorrelation} by a significant margin in the extremely inaccurate support mask scenario (mIoU of 72.5 compared to 52.7 and 45.2). Meanwhile, the robustness against image noise and distortion can also be confirmed in Figure~\ref{fig:robust_2}.

The comparisons on \textbf{parameter complexity, multi-adds and speed} are illustrated in Table~\ref{tab:complexity}. Our model strikes the best balance among complexity, performance and versatility.

\begin{table}[h]
  \centering
  \caption{Comparison on parameter complexity, multi-adds and speed. PGMA-Net-lite is a variant of PGMA-Net that without using high-order affinities.}
  \label{tab:complexity}
  \scalebox{0.8}{ 
  \begin{tabular}{c|ccccc}
    \toprule
    \multirow{2}{*}{\textbf{Method}}&\textbf{Learnable}&\textbf{Multi-adds}& \textbf{Speed}& \multirow{2}{*}{\textbf{mIoU}} & \textbf{Extra}\\
    &\textbf{parameters(M)}&\textbf{(G)}&\textbf{(fps)}&&\textbf{task}\\

    \midrule
    HSNet~\cite{min2021hypercorrelation}&2.6&\textbf{20.1}&16.0 &64.0&no\\
    DCAMA~\cite{shi2022dense}&14.2&39.8&\textbf{19.7}&64.6&no\\
    VAT~\cite{HongCNLK22}&3.2&69.0&8.1&65.3&no\\
    PGMA-Net&2.7&42.3&10.5&\textbf{74.1}&\textbf{yes}\\
    PGMA-Net-lite&\textbf{1.4}&41.8&12.0&73.1&\textbf{yes}\\
    \bottomrule
  \end{tabular}
  }
\end{table}

\begin{table*}[ht]
  \centering
  \caption{Without extra fine-tuning, the trained PGMA-Net has the ability to perform additional tasks, e.g., cross-domain FSS, bounding-box level FSS and co-segmentation tasks.}
  \label{tab:cross-domain-bbox-image} 
      \scalebox{0.8}{
      \begin{tabular}{cl|cccccc|cccccc}
              \toprule
              \multirow{2}{*}{\textbf{Task}} & \multirow{2}{*}{\textbf{Method}} & \multicolumn{6}{c|}{\textbf{1-shot}} & \multicolumn{6}{c}{\textbf{5-shot}} \\ 
              
              & & $\mathbf{5^{0}}$ & $\mathbf{5^{1}}$ & $\mathbf{5^{2}}$ & $\mathbf{5^{3}}$ & \textbf{mIoU}  & \textbf{FB-IoU} & $\mathbf{5^{0}}$ & $\mathbf{5^{1}}$ & $\mathbf{5^{2}}$ & $\mathbf{5^{3}}$ & \textbf{mIoU} & \textbf{FB-IoU} \\

              \midrule
              \multirow{4}{*}{\shortstack{cross-domain\\FSS task}}
              & HSNet~\cite{min2021hypercorrelation}   & 48.7 & 61.5 & 63.0 & 72.8 & 61.5 & - & 58.2 & 65.9 & 71.8 & 77.9 & 68.4 & - \\
              & CWT~\cite{lu2021simpler}    & 53.5 & 59.2 & 60.2 & 64.9 & 59.4 & - & 60.3 & 65.8 & 67.1 & 72.8 & 66.5 & - \\
              & CDFSS~\cite{wang2022remember}   & \textbf{57.4} & 62.2 & 68.0 & 74.8 & 65.6 & - & \textbf{65.7} & 69.2 & 70.8 & 75.0 & 70.1 & - \\
              \cline{2-14} \\[-2.0ex]
              & PGMA-Net (ours)  &  55.8 & \textbf{75.4} & \textbf{74.0} & \textbf{84.5} & \textbf{72.4} & \textbf{82.5} & 56.1 & \textbf{75.7} & \textbf{75.0} & \textbf{84.3} & \textbf{72.8} & \textbf{82.7} \\

              \midrule
              \multirow{3}{*}{\shortstack{bbox-level\\FSS task}}
              & HSNet~\cite{min2021hypercorrelation}$^{\dagger}$   & 53.4 & 64.5 & 52.7 & 51.8 & 55.6 & 70.1 & 62.6 & 69.8 & 61.1 & 59.7 & 63.3 & 75.6  \\
              & DCAMA~\cite{shi2022dense} $^{\dagger}$  & 62.2 & 70.3 & 56.3 & 56.0 & 61.2 & 73.0 & 67.2 & 72.3 & 62.4 & 61.2 & 65.8 & 76.9  \\
              \cline{2-14} \\[-2.0ex]
              & PGMA-Net (ours)   & \textbf{72.4} & \textbf{80.8} & \textbf{68.9} & \textbf{70.7} & \textbf{73.2} & \textbf{82.5} & \textbf{73.1} & \textbf{81.5} & \textbf{69.8} & \textbf{72.1} & \textbf{74.1} & \textbf{83.1}  \\

              \midrule
              \multirow{5}{*}{\shortstack{co-segmentation\\(weakly-supervised\\FSS task)}}
              & (V+S)-1~\cite{siam2020weakly} & 49.5 & 65.5 & 50.0 & 49.2 & 53.5 & 65.6 & - & - & - & - & - & - \\
              & (V+S)-2~\cite{siam2020weakly} & 42.5 & 64.8 & 48.1 & 46.5 & 50.5 & 64.1 & 45.9 & 65.7 & 48.6 & 46.6 & 51.7 & - \\
              & IMR-HSNet~\cite{wang2023iterative}   & {62.6} & {69.1} & {56.1} & {56.7} & {61.1} & - & - & {} & - & - & - & - \\
              & HSNet~\cite{min2021hypercorrelation}-RN101 & {66.2} & {69.5} & {53.9} & {56.2} & {61.5} & 72.5 & \textbf{68.9} & 71.9 & 56.3 & 57.9 & 63.7 & 73.8 \\
              \cline{2-14} \\[-2.0ex]
              & PGMA-Net (ours)  & \textbf{68.6} & \textbf{76.3} & \textbf{60.3} & \textbf{64.1} & \textbf{67.3} & \textbf{77.8} & \textbf{68.9} & \textbf{76.6} & \textbf{60.5} & \textbf{64.1} & \textbf{67.5} & \textbf{78.0}  \\
              
              \bottomrule
      \end{tabular}
      }

  \hfill

\end{table*}

\begin{table*}[ht]
  \small
  \centering
  \caption{The ablation studies to evaluate the effects of diverse interactions among priors and affinities of PGMA-Net. While lines 1-4 include only query image, which constitutes 0-shot segmentation task, lines 5-8 comprise both query and support images, thus leading to 1-shot FSS task.}
  \label{tab:ablation_implemetation}
  \scalebox{0.7}{
  \begin{tabular}{cccccccccccc|cccc}
  \toprule
  \multirow{2}{*}{\textbf{No.}}&\multirow{2}{*}{\textbf{Description}}&\multirow{2}{*}{$p_{q}^{clip}$} &\multirow{2}{*}{$M_{q}^{v}$}&\multirow{2}{*}{$A_{qq} \cdot p_{q}^{clip}$} &\multirow{2}{*}{$A_{qq} \cdot M_{q}^{v}$}&\multirow{2}{*}{$A_{qq}^{'} \cdot p_{q}^{clip}$}&\multirow{2}{*}{$A_{qq}^{'} \cdot M_{q}^{v}$}&\multirow{2}{*}{$A_{sq}^{T} \cdot M_{s}^{gt}$}&\multirow{2}{*}{ $A_{sq}^{T} \cdot p_{s}^{clip}$}&\multirow{2}{*}{$A_{sq}^{'T} \cdot M_{s}^{gt}$}&\multirow{2}{*}{$A_{sq}^{'T} \cdot p_{s}^{clip}$}& \multicolumn{3}{c}{\textbf{0/1-shot}}\\
  &&&&&&&&&&&&\textbf{mIoU}&\textbf{FB-IoU}& $\mathbf{\Delta}\textbf{(mIoU)}$ \\     
  \midrule
       1&$I_{q}$ only&$\checkmark$&&&&&&&&&&66.1&76.8&-\\
       2&$I_{q}$, w/ $A$ &$\checkmark$&&$\checkmark$&&&&&&&&69.2&79.2&3.2\\
       3&$I_{q}$, w/ $A^{'}$ &$\checkmark$&&&&$\checkmark $&&&&&&69.0&79.4&2.9\\
       4&$I_{q}$, w/ $A$ and $A^{'}$&$\checkmark$&&$\checkmark$&&$\checkmark$&&&&&&70.6&80.0&4.5\\
       \midrule
       5&$I_{q}$ and $I_{s}$ only &$\checkmark$&$\checkmark$&&&&&&&&&71.6&81.4&5.5\\
       6&$I_{q}$ and $I_{s}$, w/ $A$ &$\checkmark$&$\checkmark$&$\checkmark$&$\checkmark$&&&$\checkmark$&$\checkmark$&&&73.1&82.6&7.0\\
       7&$I_{q}$ and $I_{s}$, w/ $A^{'}$&$\checkmark$&$\checkmark$&&&$\checkmark$&$\checkmark$&&&$\checkmark$&$\checkmark$&72.6&81.9&6.5\\
       8&$I_{q}$ and $I_{s}$, w/ $A$ and $A^{'}$&$\checkmark$&$\checkmark$&$\checkmark$&$\checkmark$&$\checkmark$&$\checkmark$&$\checkmark$&$\checkmark$&$\checkmark$&$\checkmark$&74.1&83.5&8.0\\
  \bottomrule
  \end{tabular}
  }
\end{table*}

\begin{table}
  \parbox{0.4\linewidth}{
  \centering
  \caption{Ablations on channel-drop, model hierarchy, training loss and HOMAU.}
  \label{tab:ablation_on_channel-drop_and_hierarchy_and_loss}
  \scalebox{0.8}{ 
       \begin{tabular}{c|c}
            \toprule
            \textbf{Method}&\textbf{mIoU}\\
            \midrule
            PGMA-Net&74.1\\
            w/o channel-drop&72.8\\
            w/o hierarchy&71.1\\
            w/o DICE&72.2\\
            w/o CE&73.6\\
            w/o HOMAU&73.1\\
            \bottomrule
       \end{tabular}
  }}
  \hfill
  \parbox{0.5\linewidth}{
    \centering
    \caption{Ablation on switching backbone from CLIP to IN1K-pretrained.}
    \label{tab:ablation_backbone}
    \scalebox{0.85}{ 
      \begin{tabular}{c|cc}
        \toprule
        \textbf{Backbone}&\textbf{Method}&\textbf{mIoU}\\
        \midrule
        \multirow{2}{*}{CLIP-RN50}&CLIPSeg~\cite{luddecke2022image}&59.5\\
        &PGMA-Net&\textbf{74.1}\\
        \midrule
        \multirow{2}{*}{IN1K-RN50}&CLIPSeg~\cite{luddecke2022image}&39.0\\
        &PGMA-Net&\textbf{65.0}\\
        \bottomrule
      \end{tabular}
    }}
\end{table}

\subsection{Ablations}

To examine the influence of key components in our model, we conducted comprehensive ablation analysis. We utilized the CLIP-RN50 backbone on $\text{PASCAL-}5^i$ dataset~\cite{shaban2017one} in 1-shot scenario for all ablation experiments.

\textbf{Ablation on the diverse interactions between priors and affinities.} 
The PGMA-Net proposed within this study contains ten distinctive interactions, detailed in Table~\ref{tab:ablation_implemetation}: 1) Lines 1-4 investigate the scenario in which solely the query image and text are available (ZSS task), with Line 1 acting as the baseline, achieving an mIoU of 66.1. The incorporation of different affinities, including a training-free affinity (Line 2: mIoU=69.2), high-order affinity (Line 3: mIoU=69.0), and the use of both affinities (Line 4: mIoU=70.6), leads to a noticeable and consistent improvement in overall performance. 2) Within the scope of the FSS task, Lines 5-8 simultaneously integrate both query and support images, yielding superior results as articulated by the mIoU gains of $66.1\rightarrow71.6\rightarrow73.1\rightarrow72.6\rightarrow74.1$ with diverse affinities. Moreover, it becomes clear that the inclusion of affinities and the corresponding support-derived information can serve to effectively enhance performance in a complementary manner.

\textbf{Ablation on the channel-drop mechanism.} The influence of the channel-drop mechanism was investigated through Table~\ref{tab:ablation_on_channel-drop_and_hierarchy_and_loss}, showing an mIoU increase from $72.8$ to $74.1$ with the help of channel-drop mechanism.

\textbf{Ablation on the model hierarchy.} The impact of the hierarchical structure was studied using two methods: 1) using the final layer of each stage of RN50, which produced features with a length of ($1,1,1,1$). 2) using all layers within each stage, which generated features with a length of ($3,4,6,3$). Ablation study revealed that utilizing all features within each stage significantly improved performance (71.1 \text{v.s.} 74.1), as shown in Table~\ref{tab:ablation_on_channel-drop_and_hierarchy_and_loss}.

\textbf{Ablation on training loss function.} 
The default loss function of our proposed method is a weighted sum of cross-entropy loss and dice loss. An ablation study was carried out to examine the sensitivity of the two loss functions. Table~\ref{tab:ablation_on_channel-drop_and_hierarchy_and_loss} displays the results, confirming that integrating cross-entropy loss and dice loss leads to a substantial performance enhancement ($72.2\rightarrow73.6\rightarrow74.1$).

\textbf{Ablation on switching backbone from CLIP to IN1K-pretrained.} To investigate the robustness on switching backbone from CLIP to IN1K-pretrained, we removed the text branch and re-train our model with IN1K-RN50. As shown in Table~\ref{tab:ablation_backbone}, albeit not specifically designed for this setting, this simple variant of PGMA-Net demonstrates a high level of compatibility and yields comparable results to recent traditional FSS methods (our 65.0 v.s. RPMG-FSS 63.3, HPA 64.8, DCAMA 64.6, only slightly lower than newest FECANet). Also, compared to CLIPSeg ($59.5\rightarrow 39.0$), our PGMA-Net ($74.1\rightarrow65.0$) showed remarkable robustness, thus confirming the effectiveness of prior assembly from only visual cues.

\section{Conclusion}

In this paper, we proposed PGMA-Net, a class-agnostic model for few-shot segmentation by integrating textual information and a new prior-to-mask mapping. We introduced a general assemble unit (GAU) and a prior-guided mask assemble module (PGMAM) to fully exploit diverse interactions across different priors and affinities in a unified manner. We also proposed a hierarchical decoder with channel-drop strategy (HDCDM), which enables the model to perform additional challenging tasks without extra fine-tuning, thus leading an any-shot segmentation framework. Our approach achieved promising performance across various tasks, including FSS, ZSS, co-segmentation, box-level and cross-domain FSS tasks. In future work, we will focus on the N-way problem and further investigate the general few-shot segmentation task to advance the research and application in this area.

{\small
\bibliographystyle{ieeetr}
\bibliography{egbib}
}

\end{document}